\documentclass[10pt,twocolumn,letterpaper]{article}

\usepackage[pagenumbers]{cvpr} 

\usepackage[dvipsnames]{xcolor}

\definecolor{cvprblue}{rgb}{0.21,0.49,0.74}
\usepackage{algorithm}
\usepackage{algpseudocode}
\usepackage{multirow}
\usepackage[pagebackref,breaklinks,colorlinks,citecolor=cvprblue]{hyperref}

\title{Point'n Move: Interactive Scene Object Manipulation on Gaussian Splatting Radiance Fields}

\author{Jiajun Huang\\
Bournemouth University \\
{\tt\small jhuang@bournemouth.ac.uk}
\and
Hongchuan Yu\\
Bournemouth University \\
{\tt\small hyu@bournemouth.ac.uk}
}

\begin{document}

\twocolumn[{%
    \renewcommand\twocolumn[1][]{#1}%
    \maketitle
    \centering
    \includegraphics[width=1.0\linewidth]{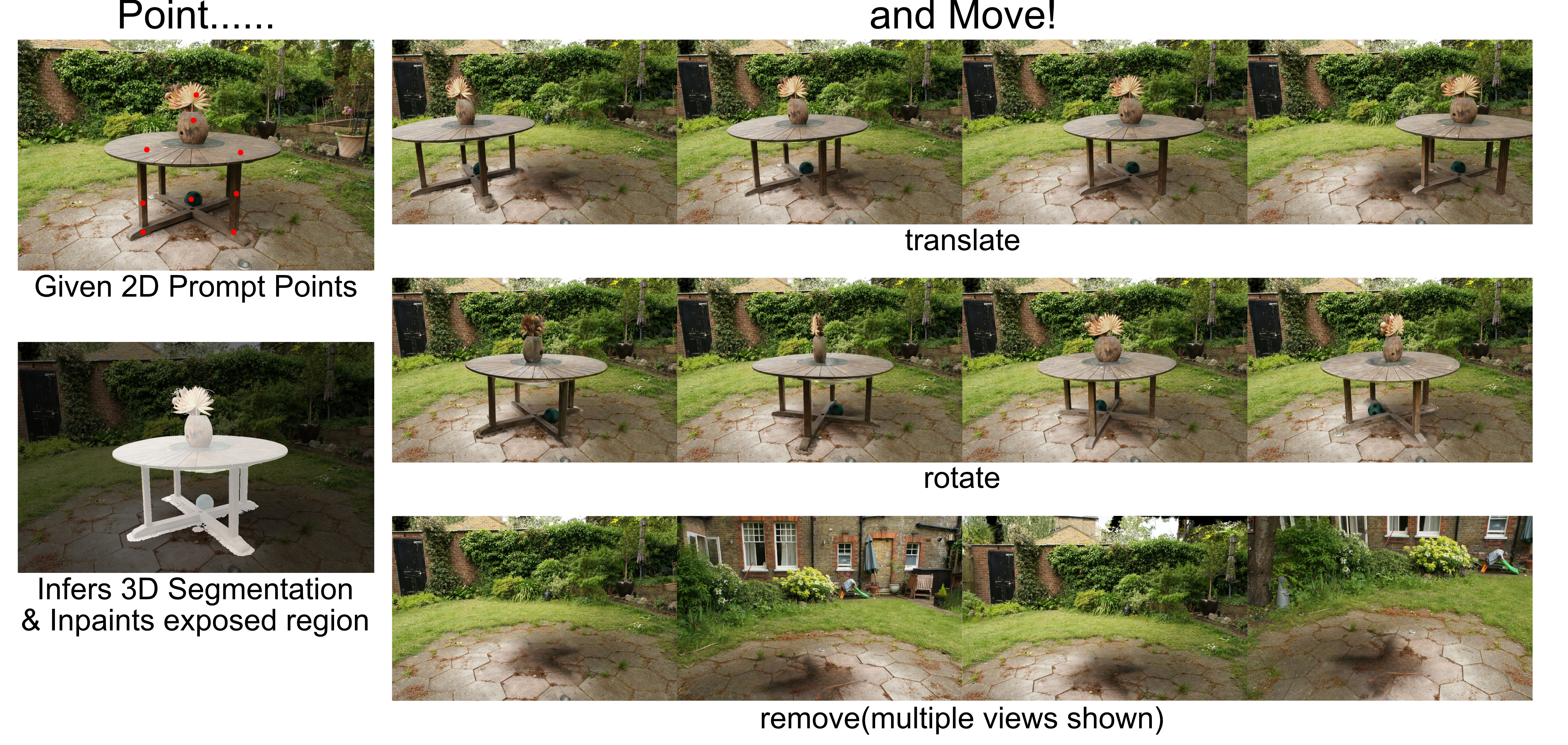}
    \captionof{figure}{
        Capability highlight of our method. The user selects an object in the scene via 2D point prompts, which our method uses to infer a 3D segmentation mask. After segmentation and exposed region inpainting, the user can freely manipulate the selected object in the scene in real-time, with all exposed regions or holes inpainted. Please refer to the supplementary material for a video demo on real-time editing.
    }
    \label{fig:highlight}
}]

\begin{abstract}

    We propose Point'n Move, a method that achieves interactive scene object manipulation with exposed region inpainting. Interactivity here further comes from intuitive object selection and real-time editing. To achieve this, we adopt Gaussian Splatting Radiance Field as the scene representation and fully leverage its explicit nature and speed advantage. Its explicit representation formulation allows us to devise a 2D prompt points to 3D mask dual-stage self-prompting segmentation algorithm, perform mask refinement and merging, minimize change as well as provide good initialization for scene inpainting and perform editing in real-time without per-editing training, all leads to superior quality and performance. We test our method by performing editing on both forward-facing and 360 scenes. We also compare our method against existing scene object removal methods, showing superior quality despite being more capable and having a speed advantage.
    
\end{abstract}    
\section{Introduction}

Radiance field-based scene representations, e.g., Neural Radiance Fields(NeRFs)\cite{nerf}, have achieved remarkable quality in capturing and representing real-life scenes. After capturing, there's a growing need to intuitively manipulate and rearrange objects within these scenes in a user-friendly manner. This enhanced editability holds immense promise for diverse applications, ranging from virtual home furnishing trials to effortless object removal from scenes and the production of VR/AR environments.

However, existing methods for editing are primarily concerned with deformation and object-centric editing, rather than freely moving or rotating objects in scenes. Methods that produce directly editable representations, such as Control-NeRF\cite{Control-NeRF}, NeuralEditor\cite{NeuralEditor} and NeuMesh\cite{NeuMesh} focus more on editing and does not provide ways to intuitively select an object from a captured scene. The real-time responsiveness poses a challenge, impeding interactive manipulation. Additionally, these methods do not inpaint the exposed surfaces or holes resulting from editing, yielding unrealistic rendering results.

In this paper, we introduce Gaussian Splatting Radiance Field(3DGS)\cite{ga3d} as the base scene representation, and present Point'n Move, a novel method for real-time interactive scene object manipulation, which includes intuitive object selection, 3D semantic segmentation, and inpainting (see \Cref{fig:highlight}). Unlike existing neural implicit representations (e.g. NeRF\cite{nerf}), 3DGS explicitly represents a radiance field using many 3D anisotropic balls, achieving rapid training and real-time rendering with impressive quality. Such point cloud-like formulation allows our method to perform segmentation and refinement from a point cloud perspective. It also serves as the foundation of our scene-content revealing pruning strategy and reprojection-based initialization, improving quality in exposed region inpainting. The speed of 3DGS also gives us a speed advantage throughout our method, especially in 3D scene inpainting. Finally, this approach enables us to directly manipulate the selected primitives in the scene rather than indirect fine-tuning, therefore achieving real-time editing.

We demonstrate our method's effectiveness by selecting and editing objects in 360 and forward-facing scenes. We also benchmark our method against existing object removal methods, achieving competitive results in quality despite being able to perform full manipulation rather than just removal and a speed advantage in scene inpainting. The code of our method will be published upon paper acceptance.

In summary, our contributions are as follows:
\begin{itemize}
  \item We propose Point'n Move, the first end-to-end method that achieves interactive scene object manipulation with exposed region inpainting on Gaussian Splatting Radiance Fields(3DGS). More concretely, our method leverages the rapid and explicit nature of 3DGS to enable intuitive selection, high-quality inpainting and real-time editing.
  \item For intuitive selection, we propose a dual-stage self-prompting mask propagation process that produces high-quality 3D semantic segmentation masks from 2D image prompt points.
  \item For high-quality exposed region inpainting, we propose a rapid inpainting procedure that minimizes unnecessary inpainting and a reprojection-based initialization scheme. Both contribute to high-quality results.
  \item Real-time editing is achieved by directly manipulating the primitives in the scene representation, all without time-consuming fine-tuning.
\end{itemize}
\section{Related Work}

\subsection{NeRF Editing}

There has been a lot of literature on scene editing based on radiance field representations, including geometry editing\cite{ornerf:30}\cite{ornerf:46}\cite{ornerf:51} and object-centric editing\cite{ornerf:34}\cite{EditNeRF}\cite{ornerf:44}. They adopt many sophisticated techniques such as fine-tuning\cite{ornerf:34}\cite{EditNeRF}, rendering rays deformation\cite{ornerf:30}\cite{ornerf:46}\cite{ornerf:51}, or editable representations\cite{ornerf:44}.

However, most works prefer neural networks to represent radiance fields, and the implicit nature of this representation usually requires time-consuming fine-tuning for edit operations. As a result, these methods cannot support practical interactive editing use cases.

To tackle this challenge, another attempt is to represent radiance fields using explicit structures, including Control-NeRF\cite{Control-NeRF} for volumetric grid, PAPR\cite{papr}, NeuralEditor\cite{NeuralEditor} and RIP-NeRF\cite{RIP-NeRF} for point clouds, NeuMesh\cite{NeuMesh} and Differentiable Blocks World\cite{DBW} for meshes. The foundation of our method, Gaussian Splatting Radiance Field\cite{ga3d}, also falls into this category.

The current body of work falls short in adequately supporting practical interactions, facing two challenges: intuitive 3D segmentation for object selection and inpainting of exposed regions or holes resulting from editing. Our method addresses these issues by leveraging the explicit representation of the Gaussian Splatting Radiance Field \cite{ga3d}. Building upon it, our method enables real-time, unrestricted editing.


\subsection{NeRF Scene Object Removal}

Scene object removal on NeRFs is another focus, aiming to select and remove objects from a trained neural radiance field. A pioneering work in object removal, SPIn-NeRF\cite{spinnerf} achieves 3D segmentation by combining interactive segmentation methods, video segmentation methods, and NeRF-based semantic mask generation method that creates object removal masks. OR-NeRF\cite{ornerf} further simplifies this process via reprojecting the masks from the segmented 2D views onto new 2D views and then applying the Segment Anything Model\cite{sam} to perform segmentation on new views. Despite achieving good results, SPIn-NeRF and OR-NeRF both need to retrain NeRF after removal, which is not suitable for real-time unrestricted editing.

Inspired by SA3D\cite{SA3D}, our approach incorporates a cross-view segmentation training process. Leveraging the explicit characteristics of the Gaussian Splatting Radiance Field\cite{ga3d}, we realize explicit 3D segmentation through weighted point clouds, providing advantages for subsequent refinement and editing.

For inpainting exposed areas, both SPIn-NeRF and OR-NeRF introduced inpainting via 2D images. The scene NeRF associated with these images is updated by fine-tuning. A patch-wise perceptual loss in the masked regions is employed for multiview consistent effect from multiple inconsistent 2D inpaintings. Apart from these two approaches, \cite{ref-guided} proposed an incremental scene inpainting and view-dependent effect estimation scheme for multiview-consistent effect. \cite{obj-removal} also proposed a novel confidence-score-based scheme to filter out view-inconsistent inpainted images, and a re-training process to update the confidence scores. Despite achieving impressive results, the complexity of these methods leads to a long training time.

Our method follows the line of SPIn-NeRF and OR-NeRF. However, we employ a simpler perceptual loss that compares the entire area inside the bounding box that bounds the mask rather than small patches inside, reducing the number of views to back-propagate through. We also tap into the superior speed of Gaussian Splatting Radiance Field, enabled by a scene content revealing pruning scheme and reprojection-based initialization process to ensure good quality.
\section{Method}

\begin{figure*}[t]
  \centering
  \includegraphics[width=\textwidth]{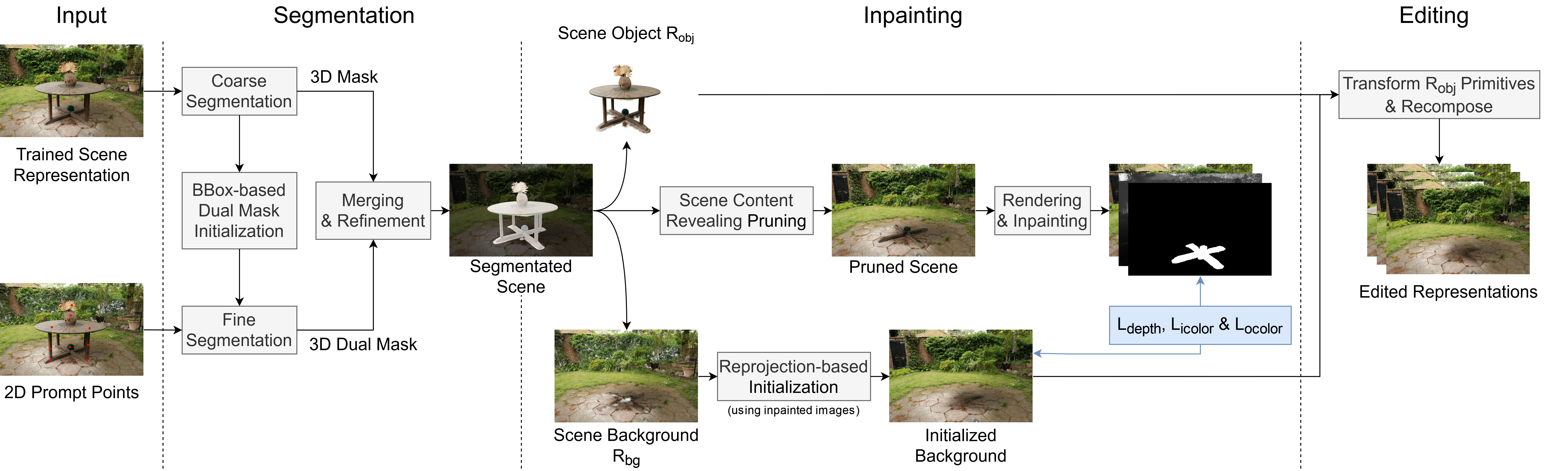}
  \caption{Overview of our method. Our method comprises of three stages: segmentation, inpainting and editing. We start with a dual-stage self-prompting cross-view mask propagation process to produce a 3D segmentation of the scene, splitting it into $R_{obj}$ and $R_{bg}$. The segmentation is then used to derive 2D scene inpaintings with a scene-content revealing pruning strategy. After using 2D inpaintings to fine-tune $R_{bg}$, $R_{obj}$ can be edited and recomposited into $R_{bg}$ in real-time.}
  \label{fig:method} 
\end{figure*}

\begin{figure*}[t]
  \centering
  \includegraphics[width=\textwidth]{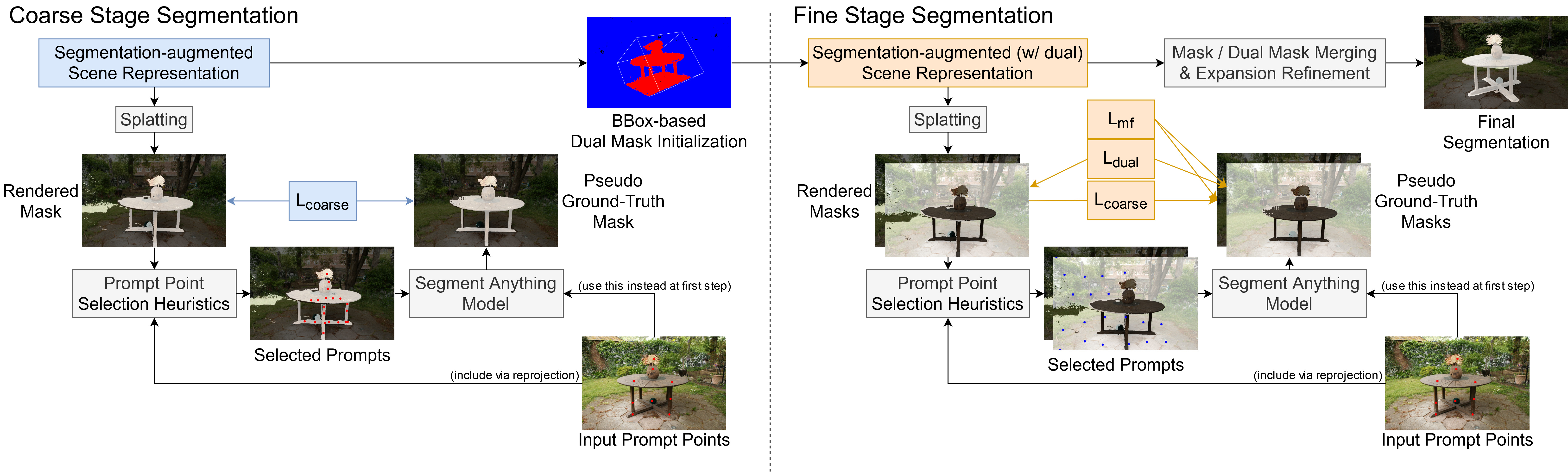}
  \caption{Detailed view of the segmentation stage. The cross-view transfer process involves selecting prompt points from rendered masks, which are used to produce pseudo-ground truths to supervise the rendering mask. The process is started using the input prompt points. After the coarse stage, a dual mask for content outside the selection and additional loss terms for training are added. Finally, the scores and dual scores are merged into the final segmentation.}
  \label{fig:method-segmentation} 
\end{figure*}

Given a trained Gaussian Splatting Radiance Field\cite{ga3d} $R$, along with a set of cameras $C$ in the scene, and some 2D point annotations $P$ that selects an object in an image rendered from $R$ at camera pose $C_0 \in C$, our method aims to achieve manipulation(translate/rotate/remove) of the object in real-time, with the newly exposed regions or holes properly inpainted. 

Our proposed approach achieves this in three steps: segmentation, inpainting and recomposition(see \Cref{fig:method}). Since our method heavily leverages its explicit formulation, we start by briefing Gaussian Splatting Radiance Fields.

\subsection{Background: Gaussian Splatting Radiance Field}

Gaussian Splatting Radiance Field\cite{ga3d}, also referred to as 3D Gaussian Splatting(3DGS), is an explicit radiance field-based scene representation that represents a radiance field using a large number of 3D anisotropic balls, each modelled using a 3D gaussian gistribution(\Cref{eq:gaussian-distribution}). More concretely, each anisotropic ball has mean $\mathcal{M} \in \mathbb{R}^3$, covariance $\Sigma$, opacity $\alpha \in \mathbb{R}$ and spherical harmonics parameters $\mathcal{C} \in \mathbb{R}^k$ ($k$ is the degrees of freedom) for modelling view-dependent colour. For regularizing optimization, the covariance matrix is further decomposed into rotation matrix $\mathbf{R}$ and scaling matrix $\mathbf{S}$ by \Cref{eq:covar-decomposition}. These matrices are further represented as quaternions $r \in \mathbb{R}^4$ and scaling factor $s \in \mathbb{R}^3$.

\begin{equation}
    \label{eq:gaussian-distribution}
    G(X)=e^{-\frac{1}{2}\mathcal{M}^T\Sigma^{-1}\mathcal{M}}
\end{equation}

\begin{equation}
    \label{eq:covar-decomposition}
    \Sigma = \mathbf{R}\mathbf{S}\mathbf{S}^T\mathbf{R}^T
\end{equation}


For this scene representation, view rendering is performed via point splatting\cite{yifan2019differentiablesplatting}. Specifically, all gaussian balls in the scene are first projected onto the 2D image plane, and their colour is computed from spherical harmonic parameters. Then, for every 16x16 pixel patch of the final image, the projected gaussians that intersect with the patch are sorted by depth. For every pixel in the patch, its colour is computed by alpha compositing the opacity and colour of all the gaussians covering this pixel by depth order, as in \Cref{eq:splatting}.

\begin{equation}
    \label{eq:splatting}
    C = \sum_{i\in N_{cov}}c_i \alpha_i \prod_{j=1}^{i-1} (1-\alpha_j)
\end{equation}
where, $N_{cov}$ represents the splats that cover this pixel, $\alpha_i$ represents the opacity of this gaussian splat multiplied by the density of the projected 2D Gaussian distribution at the location of the pixel, and $c_i$ represents the computed colour.


Gaussian Splatting Radiance Field achieves a significant advantage in training and inference speed due to its explicit approach and fast splatting rendering process. In addition to leveraging its superior speed, its point-cloud-like representation formulation enables us to perform processing from a point-cloud perspective, improving speed and quality.

\subsection{Object Segmentation}

While our method maintains the division of a scene into two parts, namely the object $R_{obj}$ and the background $R_{bg}$, we devise a dual-stage segmentation process for implementation, as shown in \Cref{fig:method-segmentation}.

\textbf{Coarse Stage Segmentation} To achieve high-quality 3D segmentation, our method builds upon the cross-view self-prompting process proposed by SA3D\cite{SA3D}. We augment the scene representation for segmentation by adding and training a segmentation score attribute to all the gaussian balls. Such score could be used to render 2D segmentation maps by treating the score as colour and rendering via the splatting process. Following SA3D, each training step starts by rendering images and segmentation masks from a chosen camera pose using the augmented representation, passing them through a heuristic algorithm to extract prompt points, and feed the prompts as well as the rendered image to Segment Anything Model(SAM)\cite{sam} to produce a more accurate pseudo ground truth mask. The pseudo-ground truth is then compared to the rendered mask for loss. The loss is presented in \Cref{eq:masking-loss}, where $M_{sam}$ is the pseudo ground truth mask, $M_r$ is the rendered mask. Herein $L_{m}$ is namely the Equation 5 in \cite{SA3D}.

\begin{equation}
    \label{eq:masking-loss}
    L_{coarse} = L_{m}(M_{sam}, M_r)
\end{equation}

\textbf{Fine Stage Segmentation} To improve segmentation quality, we expand the process by adding a fine stage and a merging process. After iterating over all the provided camera poses $C$, we enter the fine stage of segmentation. In the same way as adding segmentation scores, we further augment the representation with dual segmentation scores, aiming to capture scene contents that should not be selected. We then initialize the dual scores by computing a 3D bounding box containing all gaussians with high scores in the first stage. Every gaussian outside the bounding box has its dual score set close to one, while gaussians inside the bounding box have their dual score set to zero. We then continue the process for another full iteration, training the segmentation score and the newly added dual score, with the dual score attribute following the same self-prompting process and a new loss. The loss terms for the fine stage are presented in \Cref{eq:masking-loss-fine}. $L_{mf}$(\Cref{eq:mf}) is the loss for dual mask, where $M_{sd}$ is the pseudo ground truth dual mask, $M_{rd}$ is the rendered dual mask. We also add an additional term $L_{dual}$(\Cref{eq:ldual}) that encourages mask and dual mask to have no intersections.

\begin{equation}
    \label{eq:masking-loss-fine}
    L_{fine} = L_{coarse} + L_{mf} + L_{dual}
\end{equation}

\begin{equation}
    \label{eq:mf}
    L_{mf} = L_{m}(M_{sd}, M_{rd})
\end{equation}

\begin{equation}
    \label{eq:ldual}
    L_{dual} = L_{m}(-M_r, M_{rd}) + \lambda_{dd} L_{m}(-M_{rd}, M_r)
\end{equation}

\textbf{Merging and Refinement} Finally, we merge the scores and dual scores to create the final segmentation. We compute the 3D bounding box in the same way as in the initialization process. Then, in addition to the gaussians with a high enough segmentation score, we also include every Gaussian in the bounding box whose dual score is below a threshold into the final segmentation, effectively accepting those "rejected" by dual mask. This rejection inclusion scheme aims to cover all gaussians that are difficult to train, such as those hiding under the surface of the object. We also expand the selection by including gaussians whose mean is close enough to the already selected gaussians. 

\textbf{Remark} Thanks to the explicit representation of the Gaussian Splatting Radiance Field, we can efficiently implement the fine-stage initialization scheme, merging process, and expansion refinement. The resultant 3D segmentation mask, after merging and expansion, serves as our final output, effectively delineating the gaussians and thereby segmenting the scene into $R_{obj}$ and $R_{bg}$.

\subsection{Exposed Surface Inpainting}



After splitting, $R_{bg}$ will likely contain newly exposed surface regions or holes. For better rendering results after reintegrating the manipulated $R_{obj}$, these defects should be inpainted. We achieve this by inpainting rendered images and depth maps and then fine-tuning the scene representation with them.


\textbf{2D inpainting} To minimize the amount of inpainting needed, we remove the masked gaussians far away from the not masked gaussians in $R$. This removes all selected gaussians except those in close contact with the rest of the scene, which are very likely where the exposed regions are. These regions can be identified by rendering a segmentation mask using the pruned representation. We also add the exposed holes to the rendered mask, which is computed by including new pixels with low total opacity or colour values close to the background colour after performing the removal above. Finally, we refine the acquired inpainting masks before inpainting. (Please refer to the supplementary material for details.)

This scene content revealing pruning strategy allows us to minimize the amount of inpainting and preserve the existing scene contents as much as possible, as presented in the Pruned Scene image of \Cref{fig:method}.

We render RGB images and depth maps using this pruned representation and generate inpainting masks using the abovementioned method. The RGB images and depth maps are then inpainted by a 2D inpainter. 

\textbf{3D rapid Fine-tuning} With the inpainted images $I_i$ and depth maps $D_i$, we then perform fine-tuning to inpaint $R_{bg}$. We initialize by reprojecting the masked region of an inpainted depth map and its associated image back into the representation as new gaussians. As presented in the original 3DGS paper\cite{ga3d}, good initialization is crucial for high-quality training results.


We then train using the losses below: 

\textbf{outside mask color loss} We supervise the color of regions outside the mask via a weighted sum of pixel L1 and SSIM\cite{SSIM}:

\begin{multline}
    \label{eq:finetune-out-color}
    L_{ocolor} = (1 - \lambda_{ssim}) \text{L1}(I_i  (1 - M_i), I_r  (1 - M_i)) \\ + \lambda_{ssim} \text{SSIM}(I_i  (1 - M_i), I_r (1 - M_i))
\end{multline}
where $I_i$ and $M_i$ are the inpainted image and mask, $I_r$ is the rendered image of the representation, L1 stands for mean pixel L1 loss, $\lambda_{ssim}$ is the weight for SSIM. 

\textbf{depth loss} where for scene geometry, we employ depth map L1:

\begin{equation}
    \label{eq:finetune-depth}
    L_{idepth} = \lambda_{depth} \text{L1}(D_i, D_r)
\end{equation}
where $D_i$ is the inpainted depth map, $D_r$ is the rendered depth map, L1 stands for mean pixel L1 loss, $\lambda_{depth}$ is the weighting factor. 

\textbf{inside mask color loss} where for color inside the masked region, we adopt a perceptual color loss:

\begin{multline}
    \label{eq:finetune-in-color}
    L_{icolor} = \lambda_{lpips} \text{LPIPS}(\text{Box}(I_i, M_i), \text{Box}(I_r, M_i))
\end{multline}
where $I_i$ and $M_i$ are the inpainted image and mask, $I_r$ is the rendered colour image, $\text{Box}$ stands for a function that computes the bounding box of the mask and only keeps parts of the image in the bounding box. $\text{LPIPS}$ is the LPIPS\cite{LPIPS} perceptual metric.

Here, we employ a perceptual colour loss instead of directly comparing pixel values. This is because the 2D inpaintings are not guaranteed to be multiview consistent and a strict loss could harm quality. We also filter via the bounding box of the masked region instead of strictly in mask, this is to encourage better integration between the masked region and its surroundings.

\subsection{Editing and Recomposition}


With $R_{bg}$ inpainted, $R_{obj}$ can then be freely manipulated and composited back into $R_{bg}$ in real-time for editing. 

The selected object $R_{obj}$ could be translated and rotated by transforming the position(or mean) $\mathcal{M}$ and rotation $r$ of the underlying gaussians. Recomposition is done by adding the gaussians in $R_{obj}$ back into $R_{bg}$. Note that as these transformations could be done by multiplying transform matrices, no further training is needed. All possible surfaces and holes that could be exposed by editing have already been inpainted in the previous step.

\subsection{Implementation Details}


Both stages of the segmentation training are done using the SGD optimizer, with a learning rate of 1.0. $\lambda_{dd}$ is 0.1. For image inpainting, we employ state-of-the-art 2D image inpainting model Lama\cite{Lama} for inpainting the color image and depth map. The scene inpainting training process also uses the SGD optimizer with learning rates equal to what is specified in the original 3D Gaussian Splatting paper\cite{ga3d}. $\lambda_{ssim}$ is 0.2, $\lambda_{depth}$ is 1.0, $\lambda_{lpips}$ is 1.0. All experiments are conducted using a single A5000.
\section{Experiments}

We demonstrate the effectiveness of our method by testing it on both 360 and forward-facing scene datasets. Our method is also compared to existing scene object removal methods, reporting competitive performance in segmentation and inpainting, despite being able to do more than just removal and has a speed advantage. Finally, we conclude with ablation studies for our key contributions.

\subsection{Experiment Setups}

\textbf{Datasets} For diversity in evaluation, we test our method on the MipNeRF360\cite{mipnerf360} dataset for 360 scenes and the SPIn-NeRF\cite{spinnerf} dataset for forward-facing scenes. Both datasets consist of captured images of a scene with their associated camera parameters, which can be used to train the model to edit. As a dataset for evaluating object removal methods, the SPIn-NeRF dataset also contains scene images without the objects to remove and object segmentation masks for all captured images. For input prompt points, we make use of the point annotations from OR-NeRF\cite{ornerf}.

\textbf{Metrics} For segmentation, we report the mean accuracy and IoU(Intersection over Union) score of the rendered 2D segmentation masks across all images. For the image quality of inpainted scenes, we report the mean PSNR and LPIPS\cite{LPIPS} score between the rendered image of the inpainted scene representation and the ground truth scene image where the object is removed. We also calculate the Frechet inception distance (FID)\cite{FID} between all rendered images and ground truth images taken without the target object for image quality.

\textbf{Baselines} To evaluate the segmentation and inpainting ability of our method, we compare our approach against the state-of-the-art methods in 3D segmentation and scene object removal: SPIn-NeRF\cite{spinnerf} and OR-NeRF\cite{ornerf}. More concretely, we compare against the TensoRF variant of OR-NeRF, which is the fastest and the best overall result quality variant presented in the paper.


\subsection{Object Manipulation on 360 and Forward-facing Scenes}

\begin{figure*}[t]
  \centering
  \includegraphics[width=\textwidth]{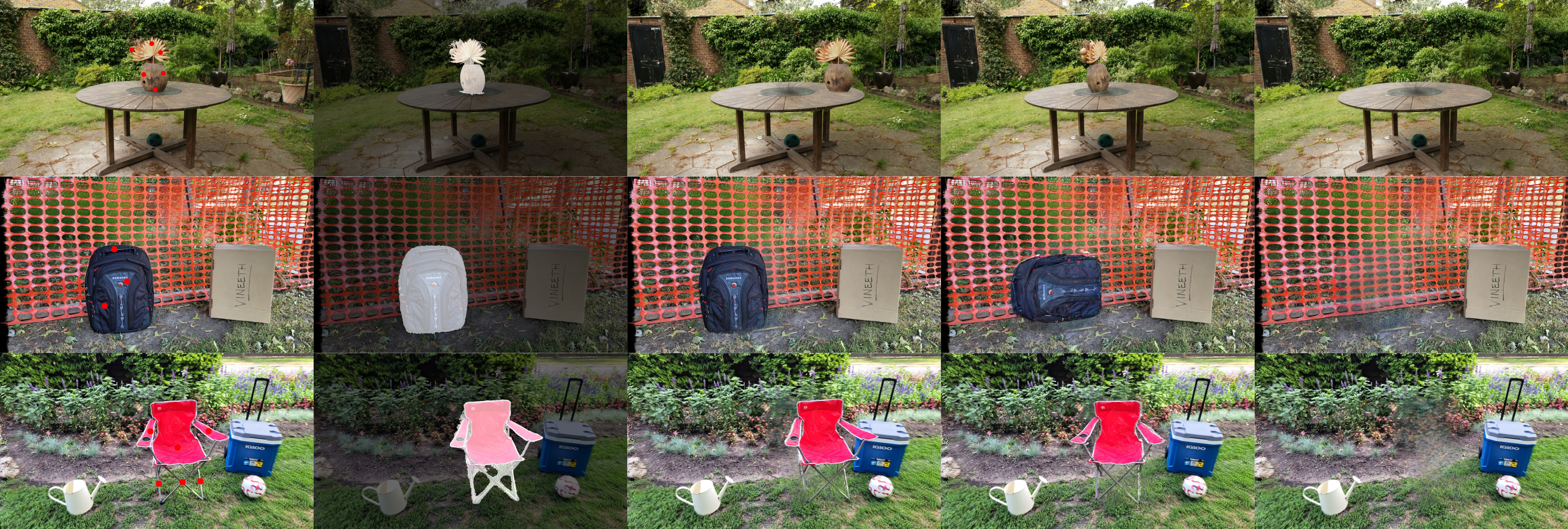}
  \caption{Input prompts, rendered segmentation mask and editing results of our method on 360 and forward-facing scenes. Note that our method gives sensible results even on scenes with extreme prompts(see last row). Please refer to the supplementary material for a video demo on real-time editing.}
  \label{fig:method-editing} 
\end{figure*}

To demonstrate the effectiveness of our approach, we perform manipulation with our method on scenes from the forward-facing SPIn-NeRF dataset and the 360 MipNeRF360 dataset. The qualitative results are presented in \Cref{fig:method-editing}. It can be noted that our method can perform high-quality selection and editing on various scenes, with exposed regions properly inpainted. Our method produces sensible results even for cases with extreme input, such as the scene at last row of \Cref{fig:method-editing}.

\subsection{Comparing Against Object Removal Methods}

\begin{figure*}[t]
  \centering
  \includegraphics[width=\textwidth]{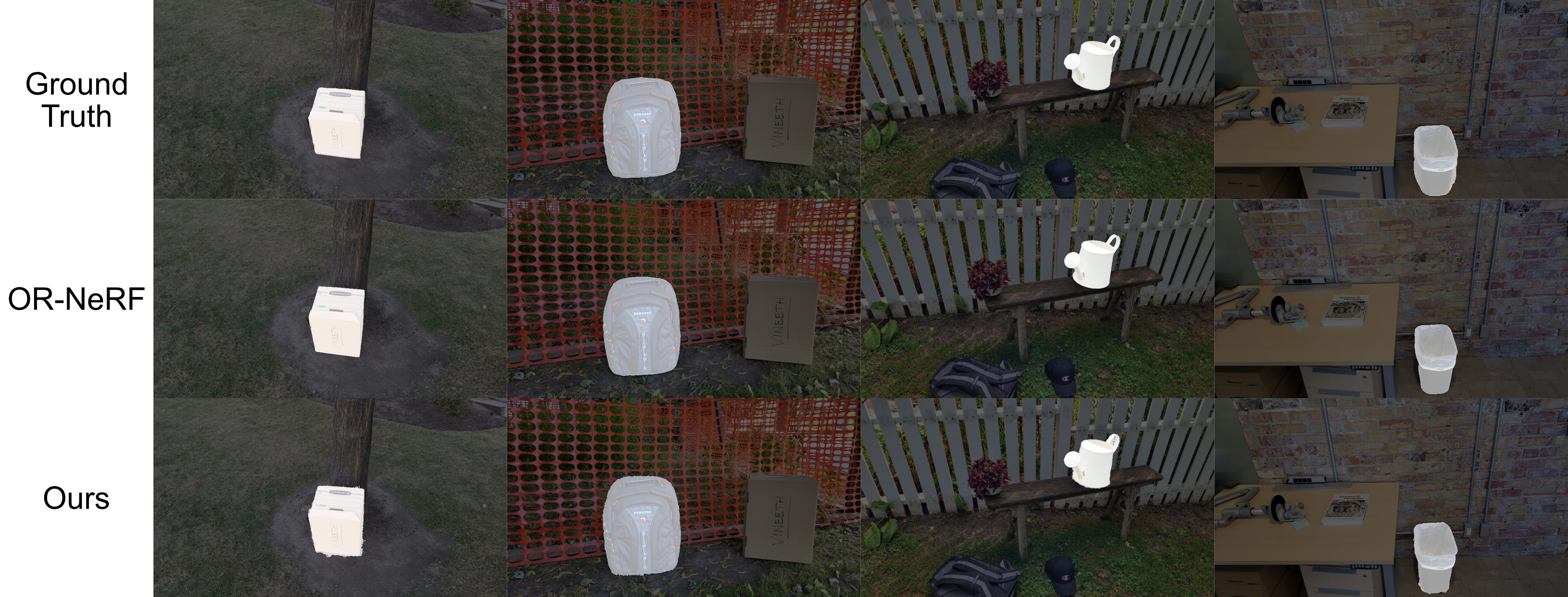}
  \caption{Qualitative comparison on segmentation against object removal methods on the SPIn-NeRF dataset. We compare by rendering our 3D segmentation mask into 2D mask images. As can be seen, our method has comparable performance against OR-NeRF(which is superior to SPIn-NeRF in segmentation), despite producing 3D segmentation instead of 2D masks on images.}
  \label{fig:cmp-segmentation} 
\end{figure*}

\begin{figure*}[t]
  \centering
  \includegraphics[width=\textwidth]{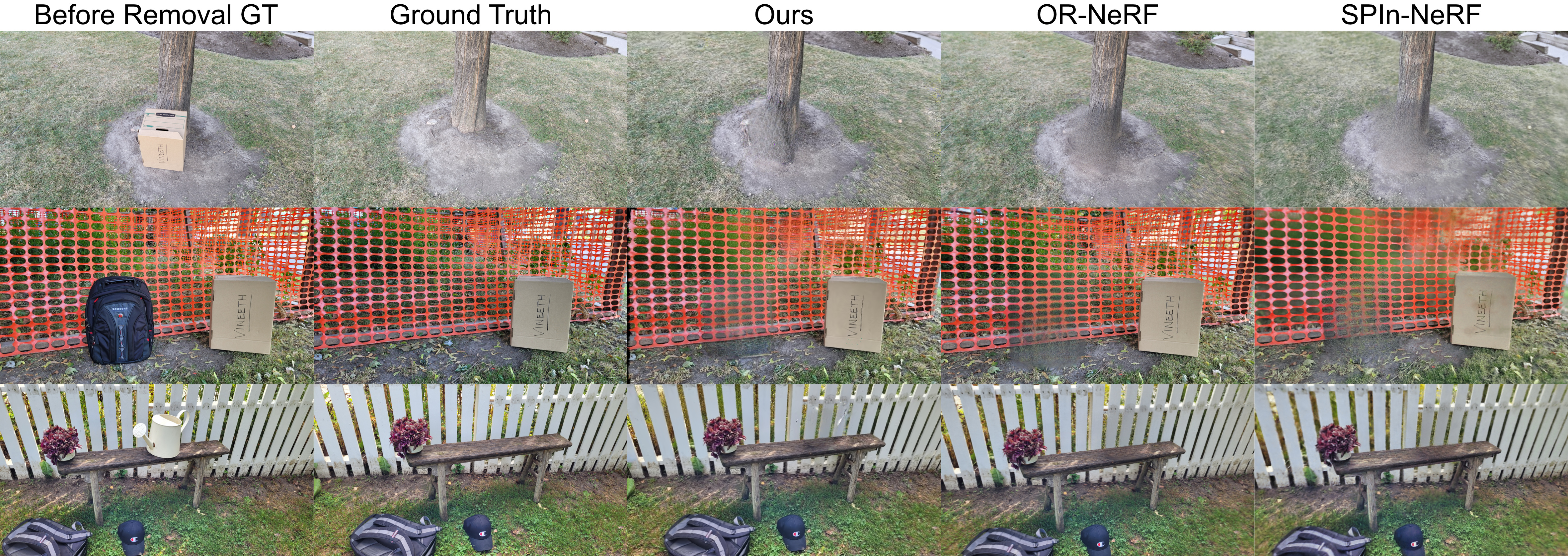}
  \caption{Qualitative comparison on exposed region inpainting on the SPIn-NeRF dataset. We present the ground truth with the object, without the object and the output of all the methods. As can be seen, our method is comparable, if not superior in terms of inpainting quality on the exposed regions.}
  \label{fig:cmp-inpainting} 
\end{figure*}

\begin{table*}[t]
    \centering
    \begin{tabular}{|c|cc|cccc|}
    \hline
    \multirow{2}{*}{method} & \multicolumn{2}{c|}{Segmentation Metrics}              & \multicolumn{4}{c|}{Inpainting Metrics}                                                                                                                   \\ \cline{2-7} 
                            & \multicolumn{1}{c|}{Acc($\uparrow$)} & IoU($\uparrow$) & \multicolumn{1}{c|}{PSNR($\uparrow$)} & \multicolumn{1}{c|}{FID($\downarrow$)} & \multicolumn{1}{c|}{LPIPS($\downarrow$)} & Inpainting Time($\downarrow$) \\ \hline
    Ours                    & \multicolumn{1}{c|}{99.51}           & 92.71           & \multicolumn{1}{c|}{\textbf{19.83}}   & \multicolumn{1}{c|}{\textbf{40.33}}    & \multicolumn{1}{c|}{\textbf{0.2684}}     & \textbf{20.41 mins}                  \\ \hline
    OR-NeRF                 & \multicolumn{1}{c|}{\textbf{99.71}}  & \textbf{95.42}  & \multicolumn{1}{c|}{13.94}            & \multicolumn{1}{c|}{49.91}             & \multicolumn{1}{c|}{0.6162}              & 168.97 mins                \\ \hline
    SPIn-NeRF               & \multicolumn{1}{c|}{98.91}           & 91.66           & \multicolumn{1}{c|}{14.83}            & \multicolumn{1}{c|}{67.26}             & \multicolumn{1}{c|}{0.6506}              & 58.25 mins                 \\ \hline
    \end{tabular}
    \caption{Quantitative results for comparing against object removal methods. The best value of each column is bolded for ease of reading. For segmentation, our method is slightly inferior compared to OR-NeRF and is on par with SPIn-NeRF. However, compared to OR-NeRF, we produce 3D segmentation rather than 2D segmentation. For scene inpainting, our method achieves superior performance both in terms of quality and speed. Please refer to the supplementary material for detailed results. 
   }
    \label{table:cmp-quantitative}
\end{table*}

We compare our model against the object removal baselines, which also support intuitive object segmentation. However, they can only remove the object, unable to move or rotate it. The experiments are conducted on the SPIn-NeRF dataset for the presence of ground-truth data.

\textbf{Segmentation Quality} For segmentation quality, the qualitative results are presented in \Cref{fig:cmp-segmentation} and the quantitative results are presented in \Cref{table:cmp-quantitative}. The table shows that our method only achieves competitive performance against OR-NeRF. We attribute this to the fact that OR-NeRF directly operates on 2D images for segmentation and creates 2D masks instead of 3D segmentation. On the other hand, our method achieves true 3D segmentation and is rendered to 2D masks for comparison. The performance of our method is on par with SPIn-NeRF\cite{spinnerf}, a 3D segmentation method.

\textbf{Scene Inpainting Quality} For inpainting  quality, the qualitative results are presented in \Cref{fig:cmp-inpainting} and the quantitative results in \Cref{table:cmp-quantitative}. As can be seen, our methods produce more plausible results compared to existing methods. We attribute this to our adopted representation's superior representation capability and our proposed initialisation and fine-tuning scheme. Our method also has a significant speed advantage in 3D scene inpainting. As shown in the last column of \Cref{table:cmp-quantitative}, our method is three times faster than the fastest baseline despite achieving superior result quality.

\subsection{Ablation Studies}

\begin{figure}[h]
  \centering
  \includegraphics[width=\linewidth]{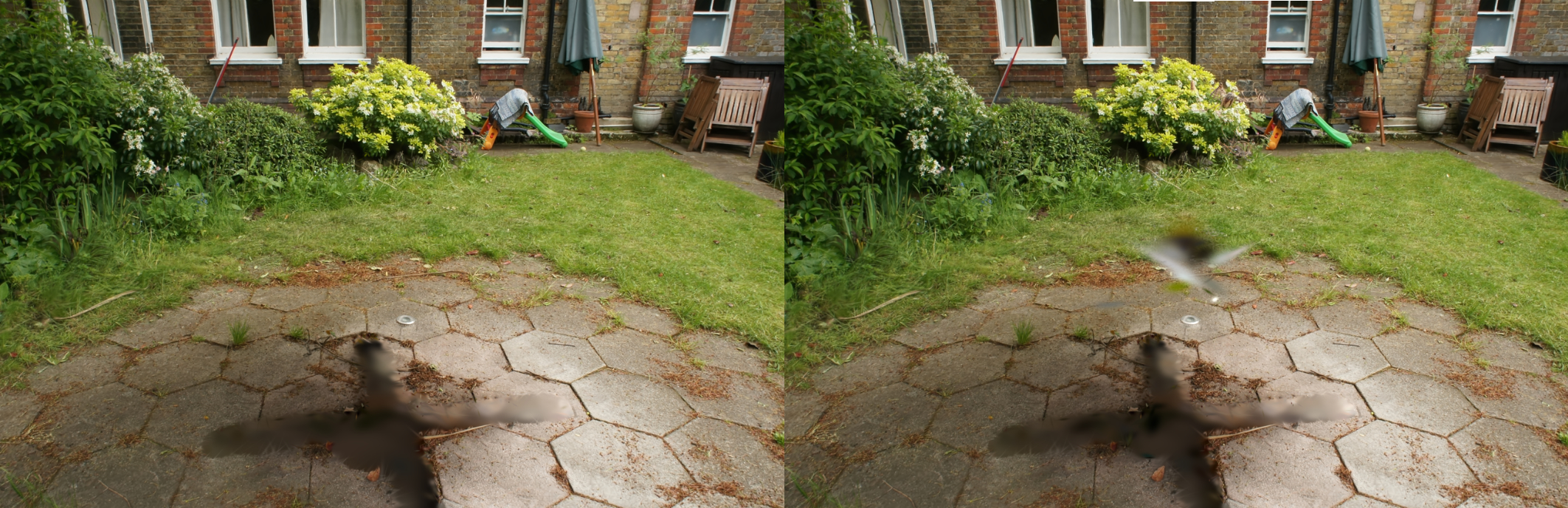}
  \caption{$R_{bg}$ part of segmentation, performed with(left) or without(right) the fine stage. As could be seen, disabling fine stage introduces floaters that should be included in $R_{obj}$.}
  \label{fig:cmp-abl-fine} 
\end{figure}

\textbf{Dual Stage Segmentation} To validate the importance of dual-stage segmentation, we disable the fine stage in our segmentation process and compare by evaluating images rendered from $R_{bg}$. As shown in \Cref{fig:cmp-abl-fine}, disabling the fine stage could leave floaters in the scene that would be fixed by the fine stage and merging process.

\begin{figure}[h]
  \centering
  \includegraphics[width=\linewidth]{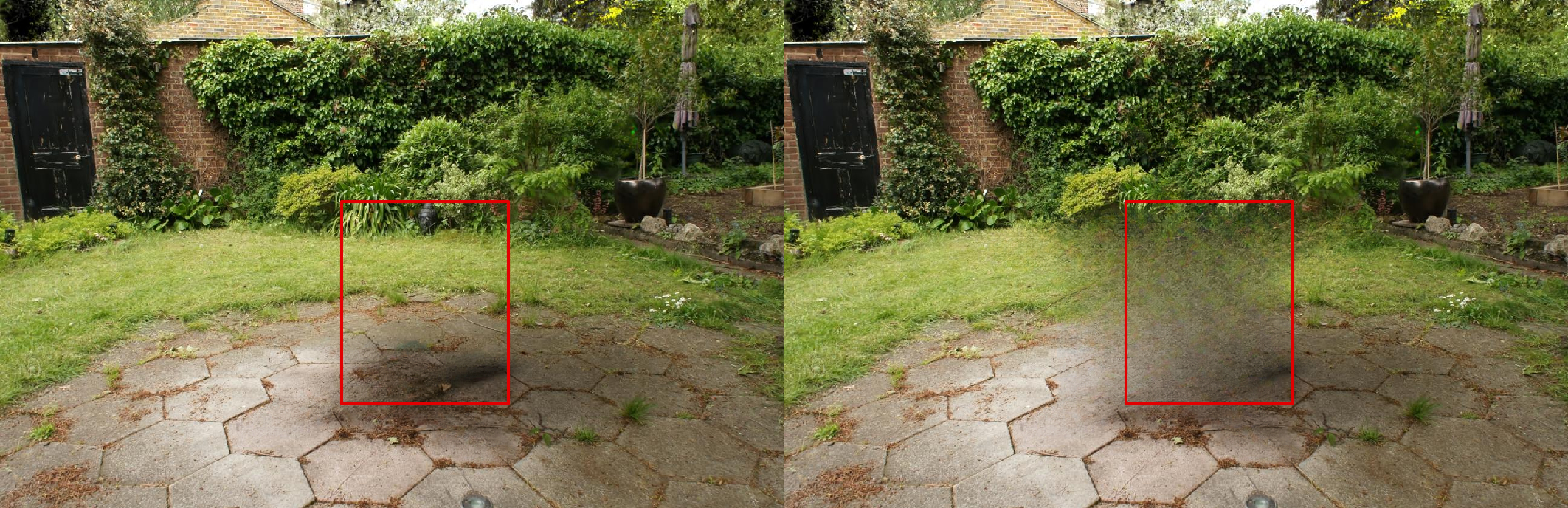}
  \caption{2D inpainting output, performed with(left) or without(right) content-revealing pruning. Note that pruning removes the blur caused by excessive inpainting in the red box region.}
  \label{fig:cmp-abl-prune} 
\end{figure}

\textbf{Content-Revealing Scene Pruning} We validate the importance of scene content revealing pruning by checking the inpainted images of the garden scene that has or has not gone through the pruning process before rendering masks and images for inpainting. As presented in \Cref{fig:cmp-abl-prune}, the pruning scheme reveals contents behind the table instead of relying on inpainting, thus improving output quality.

\begin{figure}[h]
  \centering
  \includegraphics[width=\linewidth]{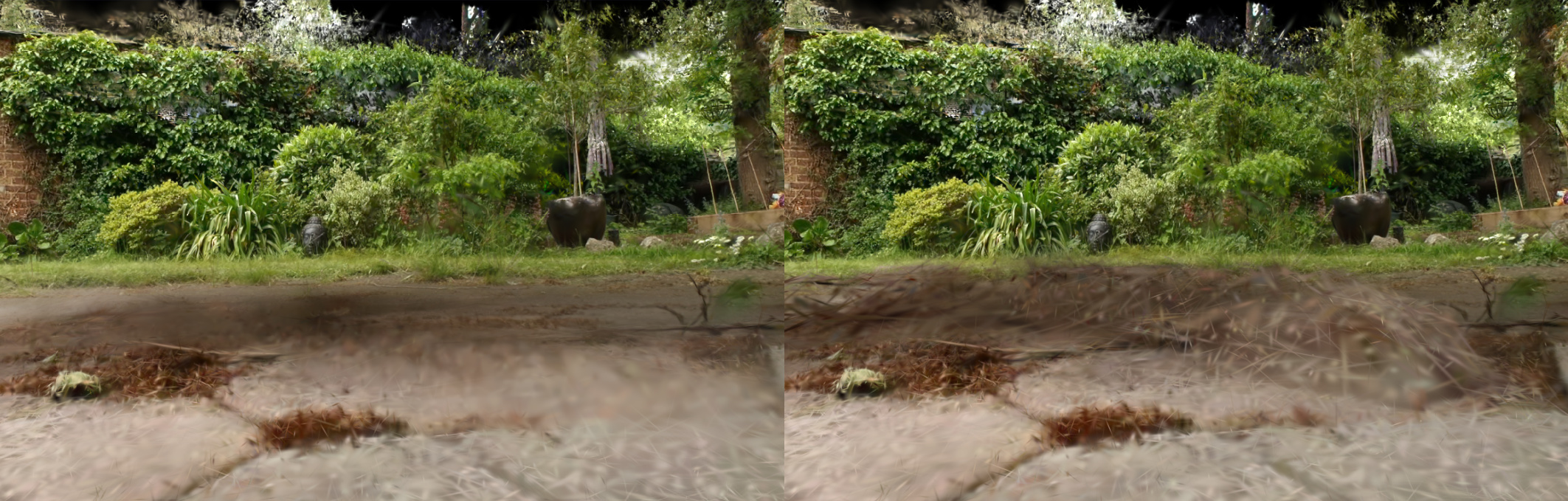}
  \caption{Close-up view of the output of scene fine-tuning, performed with(left) or without(right) reprojection-based initialization. Please note the small residual grains present on the right.}
  \label{fig:cmp-abl-no-init} 
\end{figure}

\textbf{Inpainting Optimization Initialization} We analyze the effect of initialization by disabling it and directly training the pruned scene. As shown in \Cref{fig:cmp-abl-no-init}, without proper initialization, the fine-tuning process would leave a cloud of tiny floating gaussian balls, instead of completely optimizing them away.
\section{Conclusion}

\begin{figure}
  \centering
  \includegraphics[width=\linewidth]{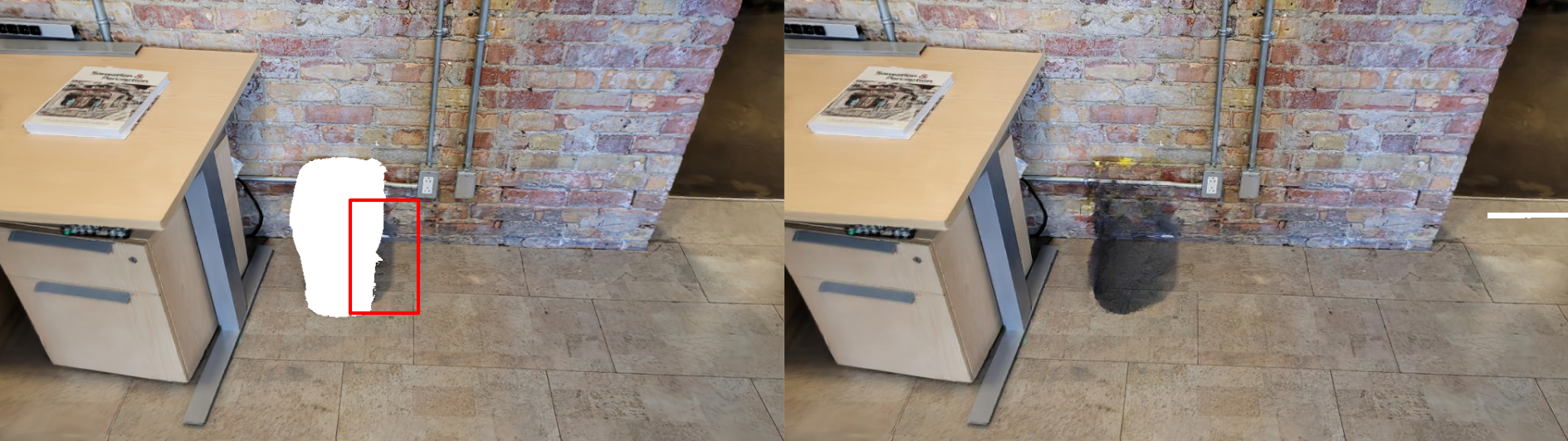}
  \caption{Artifacts caused by residual shadows misleading the inpainter. the shadows of the object(shadows in the red box on the left) could mislead the 2D inpainter, producing dark results.}
  \label{fig:limit-dark-zone} 
\end{figure}

This paper introduces Point'n Move, a methodology enabling interactive manipulation of scene objects coupled with exposed region inpainting. Leveraging the explicit, point cloud-like formulation and speed advantages offered by the Gaussian Splatting Radiance Field\cite{ga3d} as the fundamental framework of our design, our approach achieves intuitive object selection, high-fidelity exposed region inpainting, and real-time editing. These facets collectively deliver a user-friendly interactive editing experience characterized by high-quality results.

\textbf{Limitations} Currently, our method does not handle lighting or texture and focuses solely on geometry editing. Also, for some scenes, the inpaintings produced by our method are darker than expected, as presented in \Cref{fig:limit-dark-zone}. We attribute this to the precision of our segmentation as it does not include the shadow of the object, which misleads the 2D inpainter to inpaint with the shadow, making the output darker.

{
    \small
    \bibliographystyle{ieeenat_fullname}
    \bibliography{main}
}

\clearpage
\setcounter{page}{1}
\maketitlesupplementary

\section{2D Mask Refinement Algorithm}

Our 2D mask refinement algorithm(see \Cref{alg:refinement}) is inspired by the refinement procedure present in the implementation of OR-NeRF\cite{ornerf}.

In essence, our method dilates the mask, finds the contour with the largest area and takes the region inside it as the refined mask. This is to eliminate potential holes in the mask. All functions invoked in \Cref{alg:refinement} can be implemented using functions from OpenCV.

\begin{algorithm}
\caption{2D Mask Refinement Algorithm}
\label{alg:refinement}
\begin{algorithmic}
\Require 2D Inpainting Mask $m$
\Ensure Refined Inpainting Mask $ret$

\State $ m \gets \text{dilate}(m, kernel\_size=3, iteration=3) $

\State $contours \gets \text{findContours}(m)$
\State $contour \gets \text{contourWithMaxArea}(contours)$
\State $ret \gets \text{filledContour}(contour)$

\State \Return $ret$

\end{algorithmic}
\end{algorithm}

\section{Additional Per-Scene Metrics}

Quantitative results presented in \Cref{table:cmp-quantitative} is the mean across all scenes in the SPIn-NeRF\cite{spinnerf} dataset. We present the quantitative result for each individual scene here. Please see \Cref{table:cmp-segmentation-metric} for segmentation metrics and \Cref{table:cmp-inpainting-metric} for inpainting metrics.

\begin{table*}
    \centering
    \begin{tabular}{|l|c|c|c|c|c|c|c|c|c|c|}
    \hline
    metric & method & 2 & 3 & 4 & 7 & 10 & 12 & book & trash & mean \\ \hline
    \multirow{3}{*}{Acc($\uparrow$)} & Ours & \multicolumn{1}{l|}{99.80} & \multicolumn{1}{l|}{\textbf{99.79}} & \multicolumn{1}{l|}{99.74} & \multicolumn{1}{l|}{99.59} & \multicolumn{1}{l|}{99.84} & \multicolumn{1}{l|}{98.72} & \multicolumn{1}{l|}{99.01} & \multicolumn{1}{l|}{\textbf{99.56}} & \multicolumn{1}{l|}{99.51} \\
     & OR-NeRF & \textbf{99.82} & 99.73 & \textbf{99.79} & \textbf{99.81} & \textbf{99.87} & \textbf{99.30} & \textbf{99.51} & 99.51 & \textbf{99.71} \\
     & SPIn-NeRF & - & - & - & - & - & - & - & - & 98.91 \\ \hline
    \multirow{3}{*}{IoU($\uparrow$)} & Ours & \multicolumn{1}{l|}{96.16} & \multicolumn{1}{l|}{\textbf{98.09}} & \multicolumn{1}{l|}{98.15} & \multicolumn{1}{l|}{94.68} & \multicolumn{1}{l|}{94.67} & \multicolumn{1}{l|}{86.51} & \multicolumn{1}{l|}{83.45} & \multicolumn{1}{l|}{\textbf{89.99}} & \multicolumn{1}{l|}{92.71} \\
     & OR-NeRF & \textbf{96.47} & 97.48 & \textbf{98.50} & \textbf{97.43} & \textbf{95.47} & \textbf{91.73} & \textbf{88.68} & 88.68 & \textbf{95.42} \\
     & SPIn-NeRF & - & - & - & - & - & - & - & - & 91.66 \\ \hline
    \end{tabular}
    \caption{Quantitative per-scene result on segmentation quality. Results for SPIn-NeRF and OR-NeRF are taken from their original papers. SPIn-NeRF did not provide per-scene metrics; hence, they are not shown. The best value of each column is bolded for ease of reading. Our method is slightly inferior to OR-NeRF but has an advantage over SPIn-NeRF. However, we produce 3D segmentation, while OR-NeRF performs segmentation on 2D images, which is not usable for unrestricted manipulations}
    \label{table:cmp-segmentation-metric}
\end{table*}

\begin{table*}
    \centering
    \begin{tabular}{|c|c|c|c|c|c|c|c|c|c|c|}
    \hline
    metric & method & 2 & 3 & 4 & 7 & 10 & 12 & book & trash & mean \\ \hline
    \multirow{3}{*}{PSNR($\uparrow$)} & Ours & \textbf{18.48} & \textbf{18.04} & \textbf{20.88} & \textbf{21.4} & \textbf{19.75} & \textbf{16.62} & \textbf{22.28} & \textbf{21.18} & \textbf{19.83} \\
     & OR-NeRF & 15.94 & 11.42 & 13.02 & 14.37 & 12.89 & 11.40 & 15.88 & 16.63 & 13.94 \\
     & SPIn-NeRF & 16.69 & 12.08 & 14.90 & 15.34 & 12.73 & 12.39 & 17.84 & 16.70 & 14.83 \\ \hline
    \multirow{3}{*}{FID($\downarrow$)} & Ours & \textbf{53.60} & 36.39 & \textbf{51.78} & \textbf{22.48} & \textbf{21.67} & \textbf{26.23} & 81.68 & \textbf{28.84} & \textbf{40.33} \\
     & OR-NeRF & 72.10 & \textbf{34.72} & 74.04 & 38.66 & 43.89 & 38.02 & \textbf{64.91} & 32.96 & 49.91 \\
     & SPIn-NeRF & 71.75 & 68.35 & 61.10 & 43.95 & 91.73 & 50.52 & 102.71 & 47.98 & 67.26 \\ \hline
    \multirow{3}{*}{LPIPS($\downarrow$)} & Ours & \textbf{0.4544} & \textbf{0.2217} & \textbf{0.3229} & \textbf{0.2858} & \textbf{0.2264} & \textbf{0.3352} & \textbf{0.2147} & \textbf{0.2301} & \textbf{0.2864} \\
     & OR-NeRF & 0.7909 & 0.4937 & 0.6684 & 0.6445 & 0.6165 & 0.7179 & 0.5094 & 0.4882 & 0.6162 \\
     & SPIn-NeRF & 0.8489 & 0.5472 & 0.6815 & 0.6552 & 0.7003 & 0.7518 & 0.4226 & 0.5972 & 0.6506 \\ \hline
    \end{tabular}
    \caption{Quantitative per-scene result on scene inpainting quality. Results for SPIn-NeRF and OR-NeRF are taken from their original papers. The best value of each column is bolded for ease of reading. Please note our method's overall best performance across most scenes.}
    \label{table:cmp-inpainting-metric}
\end{table*}

\end{document}